\documentclass{article}

\usepackage[final]{neurips_2023}

\usepackage[utf8]{inputenc} 
\usepackage[T1]{fontenc}    
\usepackage{hyperref}       
\usepackage{url}            
\usepackage{booktabs}       
\usepackage{amsfonts}       
\usepackage{nicefrac}       
\usepackage{microtype}      
\usepackage{xcolor}         

\usepackage{amsmath,amsfonts,bm}

\def\eqref#1{equation~\ref{#1}}

\def\1{\bm{1}}

\DeclareMathAlphabet{\mathsfit}{\encodingdefault}{\sfdefault}{m}{sl}
\SetMathAlphabet{\mathsfit}{bold}{\encodingdefault}{\sfdefault}{bx}{n}

\newcommand{\E}{\mathbb{E}}

\newcommand{\R}{\mathbb{R}}

\DeclareMathOperator*{\argmax}{arg\,max}

\usepackage{microtype}
\usepackage{graphicx}
\usepackage{subfigure}
\usepackage{booktabs} 
\usepackage{multirow}

\usepackage{amsthm}
\usepackage{algorithmic}
\usepackage{algorithm}
\usepackage{wrapfig}

\usepackage{csquotes}

\usepackage{breqn}

\usepackage[createShortEnv]{proof-at-the-end}

\theoremstyle{plain}
\newtheorem{theorem}{Theorem}
\newtheorem{proposition}[theorem]{Proposition}

\theoremstyle{definition}

\theoremstyle{remark}

\usepackage{bm}
\usepackage{bbm}

\DeclareMathOperator{\indicator}{\mathbbm{1}}
\DeclareMathOperator{\N}{\mathbb{N}}

\DeclareMathOperator*{\x}{\mathbf{x}}

\DeclareMathOperator*{\m}{\mathbf{m}}

\DeclareMathOperator{\X}{\mathcal{X}}

\DeclareMathOperator{\Y}{\mathcal{Y}}
\DeclareMathOperator{\D}{\mathcal{D}}

\title{De Novo Drug Design with \textsc{Joint Transformer}s}

\author{%
  Adam Izdebski\\
  University of Warsaw \\
  \texttt{adam.izdebski@mimuw.edu.pl} \\
    \And
    Ewelina Weglarz-Tomczak \\
    NatInLab \\
    \texttt{e.weglarz.tomczak@natinlab.com} 
    \And
    Ewa Szczurek \\
    University of Warsaw \\
    \texttt{szczurek@mimuw.edu.pl}
    \And
    Jakub M. Tomczak \\
    Eindhoven University of Technology \\
    \texttt{j.m.tomczak@tue.nl}
}

\begin{document}

\maketitle

\begin{abstract}
\emph{De novo} drug design requires simultaneously generating novel molecules outside of training data and predicting their target properties, making it a hard task for generative models. To address this, we propose \textsc{Joint Transformer} that combines a Transformer decoder, Transformer encoder, and a predictor in a joint generative model with shared weights. We formulate a probabilistic black-box optimization algorithm that employs \textsc{Joint Transformer} to generate novel molecules with improved target properties and outperforms other SMILES-based optimization methods in \emph{de novo} drug design. 
\end{abstract}

\section{Introduction}
\emph{De novo} drug design is an approach to generate novel structures with desired properties from scratch. It opens the door to new classes of drugs, promising to overcome limitations of existing treatments  \citep{schneider2019automated}. While numerous breakthroughs in generative modeling and natural language processing \citep{vaswaniAttentionAllYou2017, radford2018improving} advanced the field of drug discovery, \emph{de novo} design remains a notoriously challenging task \citep{WU202118, grisoniChemicalLanguageModels2022}.

\emph{De novo} design requires to simultaneously (i) generate novel compounds, (ii) accurately predict their target properties and (iii) optimize the generation of compounds towards the desired properties \citep{brownGuacaMolBenchmarkingModels2019}. However, as the desired properties are rarely observed in the training data, there is an inherent trade-off between generation, prediction, and optimization. The more optimized towards properties from outside the training distribution the generation is, the less reliable the generation of compounds and prediction of their properties become. 

Previous work on generative models for \emph{de novo} drug design focused on each of the required components separately. Decoder-only Transformers \citep{radford2018improving} successfully generate novel and chemically plausible molecules \citep{bagalMolGPTMolecularGeneration2022}, but they have no information about the target properties. These can be fine-tuned or coupled with RL approaches, however, without yielding satisfactory results in practical regimes \citep{neil2018exploring}. Encoder-only Transformers excel at molecular property prediction tasks \citep{ross2022largescale, zhou_gao_ding_zheng_xu_wei_zhang_ke_2023}, but they lack molecule generation capabilities. Optimization of molecules is often treated as a black-box optimization (BBO) problem \citep{terayamaBlackBoxOptimization2021} and solved over a continuous latent space of a Latent Variable Model like a Variational Autoencoder \citep{kingma2013auto, rezende2014stochastic}, using an external optimization routine, e.g., Bayesian Optimization \citep{gomez-bombarelliAutomaticChemicalDesign2018, tripp2020sample}. However, posing the problem as BBO and employing an external optimization routine tend to guide a generative model far from the true data distribution of chemically plausible molecules, in regions where the generation and prediction becomes unreliable. The lack of a coherent framework that would address all the above challenges at the same time motivates the need for a joint approach.

In this paper, we propose \textsc{Joint Transformer}, a joint generative model that simultaneously generates novel examples and accurately predicts their target properties. We achieve this by combining a Transformer decoder (driving the generative performance) with a Transformer encoder and a predictor (both encouraging predictive performance). We propose to train the joint generative model with a penalized log-likelihood objective, which allows simultaneous training of the decoder, encoder, and predictor, enabling joint training and sharing all the weights. Equipped with \textsc{Joint Transformer}, we pose \emph{de novo} drug design as a probabilistic version of the BBO problem, where we aim to optimize a given objective function, like in BBO, but only in the regions of the input space, where the generative model assigns high likelihood to samples and the generative and predictive capabilities of the model remain reliable. Finally, we propose a sampling algorithm that utilizes the strong generative and predictive performance of the model to generate optimized molecules. In the experiments on GuacaMol benchmark \citep{brownGuacaMolBenchmarkingModels2019}, we show that \textsc{Joint Transformer} outperforms standard SMILES-based approaches for \textit{de novo} drug design.


\section{Methodology}\label{section:methodology}

\paragraph{Problem Statement}\label{section:problem-statement}
We define the problem of \textit{de novo} drug design as an extension of the \textit{black-box} optimization (BBO) \citep{alarieTwoDecades2021, audet2017derivative, terayamaBlackBoxOptimization2021}, in which 
we aim to find examples $\x^* \in \X$ that maximize a given \emph{objective function} $f: \X \to \R$, ${\x}^* = \argmax_{\x \in \X} f(\x)$. Additionally, we define define an example $\x \in \X$ as \emph{semantically meaningful}, if $\x$ could have been generated by the true data generating distribution $p(\x)$. This leads to the \textit{probabilistic} BBO (PBBO) defined as the problem of sampling examples $\x^* \in \X$ maximizing the objective function $f$ that could have been generated by the true underlying data distribution, i.e., they are semantically meaningful:
\begin{equation}\label{eq:problem-formulation-probabilistic}
    {\x}^* \sim p(\x \mid y_{\rm{max}}), \; \text{where} \; y_{\rm{max}} = \max_{\x \sim p(\x)} f(\x).
\end{equation}

\paragraph{Our approach}
To address the problem of learning and sampling from the conditional distribution $p(\x \mid y)$ in Eq.~\ref{eq:problem-formulation-probabilistic},~we propose a joint generative model of examples and corresponding targets. The advantage of such an approach is twofold. First, joint modeling encourages sharing the weights used for generation and prediction, making robust prediction of target values on newly generated examples feasible.
Second, the robust predictions give a good indication of whether the newly generated examples have high values of the target. 
Indeed, the joint generative model allows sampling examples $\x$ that fulfill Eq.~\ref{eq:problem-formulation-probabilistic} and satisfy the desired condition $y\geq y_c$, for every $y_c \in \R$.

The proposed joint generative model, \textsc{Joint Transformer}, $p_{\theta, \phi}(\x, y)$ combines three models: a Transformer decoder $p_{\theta}(\x)$, a Transformer encoder $\prod_{d = 1}^D p_{\theta}(x_{d} \mid \m \odot \x_{-d})$, and a predictor $p_{\theta, \phi}(y \mid \x)$. The weights $\theta$ are shared between the encoder, decoder, and predictor parts. Additionally, the predictor (used either for regression or classification) is stacked on the top of the encoder and is parametrized with weights $\phi$. The difference between the decoder and the encoder lies only in the choice of masking used within attention layers, namely, the decoder uses causal masking while the encoder applies bidirectional masking.

The rationale behind our model is the following. First, we share weights $\theta$ to entangle the generation and prediction tasks and make robust predictions of target values on newly generated examples feasible. At the same time, sharing weights has the practical advantage of a more computationally efficient model. Second, alongside the Transformer decoder, we incorporate the Transformer encoder to let the predictor learn better representations and process the input in a non-sequential manne, as a lack of bidirectional context may be harmful to predictive performance \citep{devlin2018bert}.  

In order to learn a single model that combines a Transformer encoder, a Transformer decoder, and a predictor, we propose to minimize a penalized negative log-likelihood of the joint model given by:
\begin{equation}\label{eq:penalized_ll_joint}
    \begin{split}
        \ell(\theta, \phi) = -
        \mathbb{E}_{(\x, y) \sim p(\x, y)} \bigg\{ \ln p_{\theta}(\x) +  
        \ln p_{\theta, \phi}(y \mid \x)\; +\hspace{2.5cm}\\
        \mathbb{E}_{\m \sim q(\m)}\left[\sum_{d=1}^{D} \ln p_{\theta}(x_{d} \mid {\m} \odot {\x}_{-d}) \right] \bigg\},
    \end{split}
\end{equation}
where $q(\m)$ is an arbitrary masking distribution, and the penalty (the last component of the sum) is the \textit{pseudolikelihood} \citep{wang2019bert}. Using the penalized negative log-likelihood objective in Eq. \ref{eq:penalized_ll_joint} encourages the model to simultaneously operate in two separate modes: input generation and property prediction. First, updating the decoder and learning to process the input in an autoregressive manner drives the generative performance of the model. Second, updating the encoder and learning to process the input in a bidirectional manner drives learning a meaningful representation and the predictive performance, since the predictor shares weights $\theta$ with the encoder. The training is discussed in Appendix \ref{sect:training}.

\paragraph{Probabilistic Black Box Optimization with \textsc{Joint Transformers}}\label{section:pbbo}

We define PBBO as the problem of sampling from the conditional distribution $\x \sim p(\x \mid y_c)$, where target $y_c \in \R$ is equal or close to the optimal value of the objective function $f$. However, in Proposition~\ref{thm:sampling} in Appendix \ref{section:conditional-generation} we show that sampling $\x \sim p(\x \mid y_c)$ is equivalent to the conditional generation from a joint model like \textsc{Joint Transformer}. Moreover, in Proposition~\ref{thm:sampling-guarantees} in Appendix \ref{section:conditional-generation} we show that conditional sampling is practically feasible, as long as target $y_c$ is within the support of \textsc{Joint Transformer}. In practice, to fix a feasible threshold $y_c$, one can set a sampling budget of $B \in \N$ examples and then rank examples according to $p(y \mid \x)$ choosing the best examples available. Algorithm \ref{algo:pbbo} shows how to facilitate PBBO
using \textsc{Joint Transformer}.

\begin{algorithm}
	\caption{Probabilistic Black-Box Optimization with \textsc{Joint Transformer}}
    \label{algo:pbbo}
	\begin{algorithmic}[1]
		\REQUIRE \textsc{Joint Transformer} $p_{\theta, \phi}(\x, y)$ with parameters $\theta, \phi$.\\ Threshold $y_{c} \in \Y$. Evaluation budget $I \in \N$. Sampling budget $B \in \N$.
        \ENSURE $\D_{\rm{new}} = \{(\x_i, y_i)\}_{i = 1}^{min(I, B)}$
        \STATE $b \gets 0$, $i \gets 1$  
        \WHILE{$b < B$ \AND $i \leq I$}
        \STATE Unconditionally sample $(\x_i, y_i) \sim p_{\theta, \phi}(\x, y)$
        \IF{$y_i \geq y_{c}$}
            \STATE $\D_{\rm{new}} \gets \D_{\rm{new}} \cup \; \{(\x_i, y_i)\}$
            \STATE $i \gets i + 1$
        \ENDIF
        \STATE $b \gets b + 1$
        \ENDWHILE
 \end{algorithmic} 
\end{algorithm} 

The proposed optimization procedure addresses the needs of \textit{de novo} drug design in several ways. First, it is probabilistic and therefore tailored to avoid non-realistic molecules as sampled examples. Second, it gives a guarantee of sampling improved examples. Third, it deals with the issue of a low fraction of labeled training data (low fraction of known target values for the molecules) by incorporating unsupervised training phases.

\section{Experiments} 
\paragraph{Setting} To validate the applicability of our approach, we show that \textsc{Joint Transformer} combined with the probabilistic Black Box Optimization algorithm (Alg.~\ref{algo:pbbo}) outperforms alternative methods for generating optimized molecules, as illustrated by application to a \emph{de novo} drug design task. We choose the architecture of GPT \citep{radford2018improving} for \textsc{Joint Transformer} in all tasks. The same architecture was previously utilized in MolGPT by \cite{bagalMolGPTMolecularGeneration2022}. However, the training of our model is different from \cite{bagalMolGPTMolecularGeneration2022}, and follows Alg.~\ref{alg:joint-transformer-training}. Implementation details for \textsc{Joint Transformer} are outlined in Appendix~\ref{appendix:implementation-details}. We use a \textsc{Joint Transformer} pre-trained, in an unsupervised manner, using molecules derived from the ChEMBL 24 dataset \citep{mendezChEMBLDirectDeposition2019}, following \cite{brownGuacaMolBenchmarkingModels2019} for processing and splitting the data. We additionally fine-tune \textsc{Joint Transformer} using penalized log-likelihood on randomly selected subsets of the training data ($N=1000$) with multi-property objective functions (MPO), derived from the GuacaMol benchmark  \citep{brownGuacaMolBenchmarkingModels2019}, as continuous targets in the range $[0, 1]$, namely, Perindopril MPO, Sitagliptin MPO, and Zaleplon MPO, which are also the three hardest to optimize \citep{gaoSampleEfficiencyMatters2022b}.

\paragraph{Methods} We compare the BBO with \textsc{Joint Transformer} (Alg.~\ref{algo:pbbo}) to other SMILES-based molecule optimization methods. In particular, we choose the three best-performing methods across all tasks in a benchmark comparing 25 various optimization methods \citep{gaoSampleEfficiencyMatters2022b}: SMILES GA \citep{yoshikawaPopulationbasedNovoMolecule2018}, REINVENT \citep{olivecronaMolecularNovoDesign2017} and an LSTM combined with a hill-climbing algorithm (LSTM + HC) \citep{brownGuacaMolBenchmarkingModels2019}. Additionally, we report the Dataset Best value, which is the best value of the objective function present in the dataset, as the upper bound for all screening methods, and MolPal \citep{graffAcceleratingHighthroughputVirtual2021}, which is a deep-learning-based screening method. We include in the comparison the Variational Autoencoder (VAE) \citep{kingma2013auto, rezende2014stochastic} combined with Bayesian optimization (VAE + BO) and a Junction Tree VAE \citep{jinJunctionTreeVariational2018} combined with Bayesian optimization (JT-VAE + BO). Finally, we compare our method to a standard, unconditional decoder-only Transformer model, fine-tuned on examples with corresponding objective values above a fixed threshold $y_c \in \R$ (MolGPT + fine-tune).

\begin{figure}[hbt!]
  \centering
  \includegraphics[width=\columnwidth]{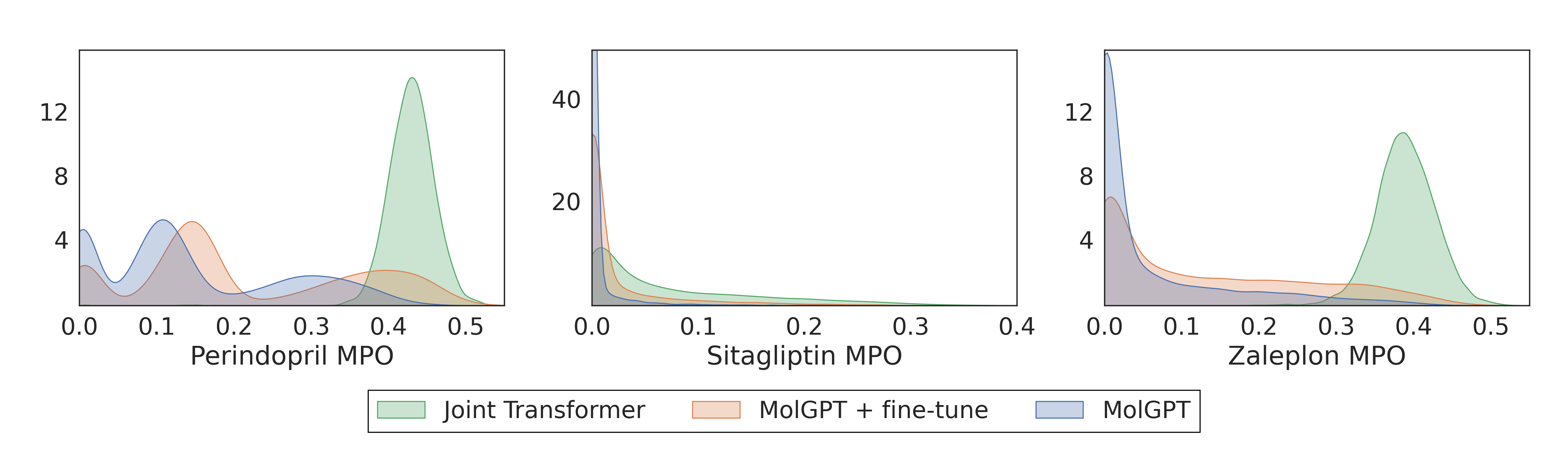}
  \vskip -18pt
  \caption{Distribution of the objective function values sampled from different models: \textsc{Joint Transformer} (green), MolGPT + fine-tune (orange), MolGPT (blue). Best viewed in color.}
\label{fig:experiment-3-distribution}
\end{figure}

\paragraph{Results} \textsc{Joint Transformer} is the only method in this experiment that generates molecules better than in the dataset across all tasks, successfully performing \emph{de novo} design (Table~\ref{table:de-novo-10^3}). The fine-tuned MolGPT is next in line in terms of performance. \textsc{Joint Transformer} generates optimized molecules for an evaluation budget as low as $137$, $452$, and $17$ evaluations, for the three optimization tasks respectively - outperforming other non-Transformer-based methods by a large margin. An additional investigation of the distribution of the objective function values for molecules sampled from the Transformer-based models as compared to the data distribution (Figure~\ref{fig:experiment-3-distribution}), shows that \textsc{Joint Transformer} significantly alters the distribution of the sampled molecules towards optimal objective values, as opposed to MolGPT that only slightly skews the initial data distribution.

\begin{table*}[h]
\caption{The highest value of the objective function across all generated molecules (Top1) within $10^3$ evaluations. 
Mean and standard deviation across three independent data splits.}
\label{table:de-novo-10^3}
\vskip 0.15in
\begin{center}
\begin{small}
\begin{sc}
\begin{tabular}{@{}lccc@{}}
\toprule
Method & Perindopril MPO & Sitagliptin MPO & Zaleplon MPO \\
\midrule
Dataset Best & $0.53 \pm 0.00$ & $0.40 \pm 0.02$ & $0.50 \pm 0.01$ \\
MolPal & $0.49 \pm 0.01$  & $0.05 \pm 0.01$ & $0.16 \pm 0.09$ \\
SMILES GA & $0.44 \pm 0.01$ & $0.21 \pm 0.10$ & $0.28 \pm 0.08$ \\
Reinvent SMILES & $0.45 \pm 0.01$ & $0.02 \pm 0.01$ & $0.27 \pm 0.03$ \\ 
LSTM + HC & $0.47 \pm 0.01$ & $0.02 \pm 0.02$ & $0.14 \pm 0.03 $ \\
VAE + BO & $0.45 \pm 0.01$ & $0.03 \pm 0.01$ & $0.04 \pm 0.04$   \\
JT-VAE + BO & $0.43 \pm 0.01$ & $0.05 \pm 0.03$ & $0.13 \pm 0.11 $ \\
MolGPT + fine-tune & $0.54 \pm 0.03$ & $0.40 \pm 0.03$ & $0.51 \pm 0.00$ \\
\textsc{Joint Transformer} (ours) & $\mathbf{0.55 \pm 0.01}$ & $\mathbf{0.43 \pm 0.01}$ & $\mathbf{0.55 \pm 0.02}$  \\
\bottomrule
\end{tabular}
\end{sc}
\end{small}
\end{center}
\vskip -0.1in
\end{table*} 

\section{Conclusion}

In this paper, we formulated the problem of de novo drug design as an instance of probabilistic BBO (Section~\ref{section:problem-statement}). We proposed a general-purpose sampling algorithm that performs probabilistic BBO with any joint generative model (Section~\ref{section:pbbo}), with theoretical guarantees on the expected runtime as the function of the training data (Section~\ref{section:conditional-generation}). Finally, we proposed a joint generative model, called \textsc{Joint Transformer}, that combines a Transformer decoder, a Transformer encoder, and a predictor in a single model with shared parameters, which is jointly trained with a penalized log-likelihood objective (Eq~\ref{eq:penalized_ll_joint}). We empirically showed that \textsc{Joint Transformer} successfully outperforms state-of-the-art approaches to \emph{de novo} drug design.


\newpage

\bibliography{neurips_2023}
\bibliographystyle{iclr2024_conference}


\newpage
\appendix

\section{Author contribution}

\textit{Anonymized for the submission} 
\textbf{AI} - devised the project, designed and performed all experiments, the lead in writing the manuscript

\textbf{EWT} - data preparation (Screening experiment), feedback on writing

\textbf{ES} - supervision, the lead in writing the manuscript

\textbf{JT} - devised the project, supervision, designed experiments and performed "Screening" experiment, the lead in writing the manuscript

\section{Training}\label{sect:training}
Training a joint generative model was previously shown to result in a good generator together with a poor predictor \citep{lasserre2006principled, nalisnick2019hybrid}. Moreover,  storing the gradients for all the summands of the objective (Eq.~\ref{eq:penalized_ll_joint}) is a significant overhead in memory requirements as compared to decoder and encoder-only Transformers. To overcome both issues, we propose a practical training procedure for \textsc{Joint Transformer} (Alg.~\ref{alg:joint-transformer-training}) that randomly switches between the input generation and the property prediction (and encoder training) tasks with a hyperparameter $p_{\rm{task}} \in [0, 1]$. 
 
The \textsc{Joint Transformer} can be trained in an unsupervised, semi-supervised or supervised setting. Depending whether a target $y \in \Y$ is sampled from the dataset $\D$ or is not available (Step~\ref{alg:step-sample}, Alg.~\ref{alg:joint-transformer-training}), one can include the prediction loss $\ln p_{\theta, \phi}(y \mid \x)$ in the penalized log-likelihood objective $\ell$  (Step~\ref{alg:step-bidirectional-update}, Alg.~\ref{alg:joint-transformer-training}), resulting in a supervised setting, or set the prediction loss to zero, resulting in an unsupervised setting. For the training data where only a small proportion of samples have accompanying target values, we split the training procedure of the \textsc{Joint Transformer} into first training the model in an unsupervised manner (Alg.~\ref{alg:joint-transformer-training-unsupervised}), then fine-tuning it with supervised data (Alg.~\ref{alg:joint-transformer-training}). 

\begin{algorithm}[H]
	\caption{Unsupervised training of \textsc{Joint Transformer}}
 \label{alg:joint-transformer-training-unsupervised}
	\begin{algorithmic}[1]
		\REQUIRE A dataset $\D = \{\x_n\}_{n=1}^{N}$. \textsc{Joint Transformer} $p_{\theta, \phi}(\x, y)$ with parameters $\theta, \phi$ containing a decoder $p_\theta(\x)$, encoder $\prod_{d=1}^D p_\theta(x_d \mid {\m} \odot {\x}_{-d})$.\\ Task probability $p_{\rm{task}} \in [0, 1]$ and a masking distribution $q(\m)$.
        \WHILE{a stopping criterion is not met}
        \STATE\label{alg:step-sample-unsupervised} Uniformly sample $\x$ from the dataset $\D$
        \STATE Sample an indicator $u \sim \textsc{Bernoulli}(p_{\textit{task}})$
        \IF{$u = 0$} 
            \STATE Sample mask $\m \sim q(\m)$
            \STATE\label{alg:step-bidirectional-update-unsupervised} Calculate loss $\ell(\theta, \phi) = -\sum_{d=1}^{D} \ln p_{\theta}(x_{d} \mid {\m} \odot {{\x}_{-d}})$ 
        \ELSE
            \STATE Set mask to the causal mask
            \STATE Calculate loss $\ell(\theta, \phi) = - \ln p_{\theta}(\mathbf{x})$
        \ENDIF
        \STATE Update parameters $\theta, \phi$ using an optimizer w.r.t.\ loss $\ell$
        \ENDWHILE
	\end{algorithmic} 
\end{algorithm} 

\begin{algorithm}[H]
	\caption{Training of \textsc{Joint Transformer}}
 \label{alg:joint-transformer-training}
	\begin{algorithmic}[1]
		\REQUIRE A dataset $\D = \{(\x_n, y_n)\}_{n=1}^{N}$. \textsc{Joint Transformer} $p_{\theta, \phi}(\x, y)$ with parameters $\theta, \phi$ containing a decoder $p_\theta(\x)$, encoder $\prod_{d=1}^D p_\theta(x_d \mid {\m} \odot {\x}_{-d})$ and a predictor $p_{\theta, \phi}(y \mid \x)$.\\ Task probability $p_{\rm{task}} \in [0, 1]$ and a masking distribution $q(\m)$.
        \WHILE{a stopping criterion is not met}
        \STATE\label{alg:step-sample} Uniformly sample $(\x, y)$ from the dataset $\D$
        \STATE Sample an indicator $u \sim \textsc{Bernoulli}(p_{\textit{task}})$
        \IF{$u = 0$} 
            \STATE Sample mask $\m \sim q(\m)$
            \STATE\label{alg:step-bidirectional-update} Calculate loss $\ell(\theta, \phi) = -\sum_{d=1}^{D} \ln p_{\theta}(x_{d} \mid {\m} \odot {{\x}_{-d}}) - \ln p_{\theta, \phi}(y \mid \x)$ 
        \ELSE
            \STATE Set mask to the causal mask
            \STATE Calculate loss $\ell(\theta, \phi) = - \ln p_{\theta}(\mathbf{x})$
        \ENDIF
        \STATE Update parameters $\theta, \phi$ using an optimizer w.r.t.\ loss $\ell$
        \ENDWHILE
	\end{algorithmic} 
\end{algorithm}

\section{Sampling}\label{section:sampling}
\subsection{Unconditional Generation}\label{section:unconditional-generation}
In the unconditional generation task, we sample from \textsc{Joint Transformer} in a two-step manner that results in an unconditional sample $(\x, y) \sim p_{\theta, \phi}(\x, y)$. First, since the decoder part $p_\theta(\x)$ does not depend on parameters $\phi$ and it properly defines an ARM, we sample $\x \sim p_\theta(\x)$. Next, we sample a target $y$ from the predictive distribution $y \sim p_{\theta, \phi}(y \mid \x)$. The key feature that allows for successful sampling from the joint model is the ability of the \textsc{Joint Transformer} to simultaneously operate in two separate modes, namely generate novel examples and predict their target values, which is directly encouraged by training with the penalized log-likelihood objective in Eq.~\ref{eq:penalized_ll_joint}. 

\subsection{Conditional Generation}\label{section:conditional-generation}
In the conditional generation task, given a condition $Y \subseteq \Y$, we sample from \textsc{Joint Transformer} $p_{\theta, \phi}(\x, y)$ to obtain a conditional sample $(\x, y) \sim p_{\theta, \phi}(\x, y)$, such that $y \in Y$. \textsc{Joint Transformer} generates conditional samples by first sampling $(\x, y) \sim p_{\theta, \phi}(\x, y)$ in the above described unconditional way and then accepting the sample if $y \in Y$. In practical applications, due to a finite runtime, we sample a batch of $B$ tuples $(\x, y) \sim p_{\theta, \phi}(\x, y)$ and choose $(\x, y)$ with $y$ `closest' to $Y$. Proposition~\ref{thm:sampling} shows that, despite its conceptual simplicity, the described conditional generation procedure is equivalent to directly sampling from the conditional distribution $p(\x \mid y)$. Moreover, Proposition~\ref{thm:sampling-guarantees} shows conditions under which conditional generation enjoys a finite expected runtime. 
\begin{proposition}\label{thm:sampling}
    Let $p(\x, y)$ be a joint probability distribution over $\X \times \Y$. Let $y_{c} \in \Y$ be such that $p(y_{c}) > 0$. Then 
    \[
    p(\x \mid y_{c}) \propto \indicator_{\{y = y_{c}\}}(y)p(y \mid \x)p(\x).
    \]
\end{proposition}
\begin{proof}
     Assume that $p(\x, y)$ is a joint probability distribution over $\X \times \Y$. Choose $y_{\max} \in Y$ to be such that $p(y \geq y_{c}) > 0$. Then a simple application of Bayes rule yields
    \begin{align}
        p(\x \mid \{y \geq y_{c}\}) = \frac{p(\x, \{y \geq y_{c}\})}{p(\{y \geq y_{c}\})} = \frac{\indicator_{\{y \geq y_{c}\}}(y)p(y \mid \x)p(\x)}{p(\{y \geq y_{c}\})}.
    \end{align}
    Since $p(\{y \geq y_{c}\}) > 0$ and it does not depend on $\x$, we have that \[
    p(\x \mid \{y \geq y_{c}\}) \propto \indicator_{\{y \geq y_{c}\}}(y)p(y \mid \x)p(\x).
    \]
\end{proof}

\begin{proposition}\label{thm:sampling-guarantees}
    Let $p(y)$ be a probability distribution over $\Y$ with a corresponding cumulative distribution function $F$. Let target $y_c \in \Y$ be such that $p(y_c) > 0$ and let $p$ be the probability of sampling a target $y \sim p(y)$ such that $y > y_c$. The expected number of trials $N$ until obtaining a sample $y \sim p(y)$ such that $y > y_c$ is equal to $1/p$.
\end{proposition}
\begin{proof}
    Let $p(y)$ be a probability distribution over $\Y$ with a corresponding cumulative distribution function $F$. Let $y_c \in \Y$ be such that $p(y_c) > 0$. Define r.v.\ $N$ as the number of trials until obtaining a sample $y > y_c$, where $y$ is distributed as $p(y)$. For each $n \in \N$, the distribution of $N$ is given by
    \[
    P(N = n) = (1 - p)^{n-1}p,
    \]
    where $p = 1 - F(y \leq y_c)$. Hence, the number of trials $N$ follows a geometric distribution with an expected value equal to $\E[N] = 1 / p$.
\end{proof} 

Despite its simplicity, the conditional generation of \textsc{Joint Transformer} has the advantage of the predictor $p_{\theta, \phi}(y \mid \x)$, as it is defined in the input space $\X$, indicating whether the newly generated example enjoys the desired target value. This is in contrast to methods based on LSO and diffusion models, see \citep{gomez-bombarelliAutomaticChemicalDesign2018, hoogeboomEquivariantDiffusionMolecule2022}. 

\section{Additional Experiments}\label{appendix:additional-experiments}

\subsection{Molecule Generation}\label{appendix:additional-experiments-molgen}

\paragraph{Task} In the molecule generation task, the goal is to generate valid and novel molecules that follow the chemical distribution of the training data. Following \citet{brownGuacaMolBenchmarkingModels2019}, we evaluate all molecule generation methods on five metrics: validity, a fraction of the generated molecules that are correspond to a valid SMILES string; uniqueness, a fraction of the generated molecules that are unique; novelty, a fraction of the generated molecules that are not present in the training data; KL Divergence, a measure of similarity of the generated molecules to the training set with respect to selected chemical properties \citep{brownGuacaMolBenchmarkingModels2019}, as well as Fr\'echet ChemNet Distance (FCD; \citep{preuerFrEchetChemNet2018}), a general measure of similarity of the generated molecules to the training set.

\paragraph{Baselines} As baselines, we select well-established molecule generation models based on SMILES representation \citep{weiningerSMILESChemicalLanguage1988}:  LSTM~\citep{ertlSilicoGenerationNovel2018}, VAE \citep{kingma2013auto, rezende2014stochastic} and AAE \citep{kadurinCornucopiaMeaningfulLeads2016}. Additionally, we consider graph-based models: Junction Tree VAE \citep{jinJunctionTreeVariational2018}, MoLeR \citep{maziarzLearningExtendMolecular2022} and MAGNet \citep{hetzelMAGNetMotifAgnosticGeneration2023}. Finally, we include MolGPT \citep{bagalMolGPTMolecularGeneration2022}, which is a Transformer-based model and the backbone for the \textsc{Joint Transformer}, sharing the same architecture, but trained differently. 

\paragraph{Results} In the molecule generation task, \textsc{Joint Transformer} successfully generates valid, unique and novel molecules~(Tab.~\ref{table:molecule-generation}). Moreover, \textsc{Joint Transformer} generates molecules with properties that closely follow the training set distribution, making the newly generated molecules realistic and physio-chemically plausible, as measured by KL Divergence and FCD. Compared to the backbone MolGPT model, \textsc{Joint Transformer} achieves identical performance, showing that the modified training procedure does not hurt the generative functionality of the model. From the generative modeling perspective, this result is counterintuitive, as we can include the reconstruction task to the training procedure of the \textsc{Joint Transformer}, without sacrificing its generative performance. 

Overall, none of the molecule generation methods achieves best performance across all metrics. Graph-based methods outperform others on validity, as they generate always valid molecules by design. However, the improvement of $3\%$ as compared to Transformer-based models (\textsc{Joint Transformer} and MolGPT) is negligible. Additionally, it comes at the expense of generating molecules with decreased (from $12\%$ to $19\%$) values of the KL Divergence and FCD metrics. On the other hand, LSTM  achieves top performance on KL Divergence and FCD metrics, slightly (1\% and 3\%, respectively) outperforming Transformer-based methods, but falls behind in the validity of the generated molecules. All methods successfully generate unique and novel molecules. Overall, \textsc{Joint Transformer} strikes a good balance between graph-based and SMILES-based LSTM, making it a viable choice for a go-to molecule generation model. 

\begin{table*}[h]
\caption{Molecule Generation Task. \textsc{Joint Transformer} (JT) matches state-of-the-art performance of different molecule generation methods. Training the \textsc{Joint Transformer} model on generation and reconstruction tasks simultaneously does not hurt the generation performance of the model.}
\label{table:molecule-generation}
\vskip 0.15in
\begin{center}
\begin{small}
\begin{sc}
\resizebox{\columnwidth}{!}{%
\begin{tabular}{@{}lcccccc@{}}
\toprule
Model & Size & Validity $(\uparrow)$ & Uniqueness $(\uparrow)$ & Novelty $(\uparrow)$ & FCD $(\uparrow)$ & KL Div $(\uparrow)$ \\
\midrule
LSTM & - & 0.96 & \textbf{1.0} & 0.91 & \textbf{0.91} & \textbf{0.99} \\
VAE & - & 0.87 & \textbf{1.0} & 0.97 & 0.86 & 0.98 \\
AAE & - & 0.82 & \textbf{1.0} & \textbf{1.0} & 0.53 & 0.89 \\
JT-VAE & - & \textbf{1.0} & n/a & n/a & 0.76 & 0.94 \\
MAGNet & 6.9M & n/a & n/a & n/a & 0.73 & 0.92 \\
MoLeR & - & \textbf{1.0} & 0.99 & 0.97 & 0.78 & 0.98 \\
MolGPT & 6M & 0.98 & \textbf{1.0} & \textbf{1.0} & \textbf{0.91} & \textbf{0.99} \\
MolGPT (ours) & 6M & $0.97$ & $\mathbf{1.0}$ & $0.97$ & $0.89$ & $0.98$ \\
\midrule
JT (ours) & 6M & $0.97$ & $\mathbf{1.0}$ & $0.98$ & $0.89$ & $\mathbf{0.99}$ \\ 
JT (ours) & 50M& $0.98$ & $\mathbf{1.0}$ & $0.95$ & $0.90$ & $\mathbf{0.99}$  \\ 
\bottomrule
\end{tabular}}
\end{sc}
\end{small}
\end{center}
\vskip -0.1in
\end{table*}

\subsection{Unconditional Generation}
Moreover, the jointly trained predictor $q_{\theta, \phi}(y \mid \x)$ of the \textsc{Joint Transformer} generalizes well to data generated with the model $p_\theta(\x)$. In particular, the prediction error, as measured by mean absolute error, of the \textsc{Joint Transformer} fine-tuned on three properties from the Guacamol task \citep{brownGuacaMolBenchmarkingModels2019} do not change between the test set and newly generated data (Table~\ref{table:generalization-predictor}). This shows good generalization performance of \textsc{Joint Transformer}. 

\begin{table*}[h]
\caption{Mean absolute prediction error (MAE) for the predictor on three property prediction tasks on test and generated data. Mean and standard deviation across independent runs.}
\label{table:generalization-predictor}
\vskip 0.15in
\begin{center}
\begin{small}
\begin{sc}
\begin{tabular}{@{}lcccc@{}}
\toprule
Method & Data & Perindopril MPO & Sitagliptin MPO & Zaleplon MPO \\
\midrule

\multirow{2}[1]{*}{\textsc{Joint Transformer}} 
 & test & $0.014 \pm 0.004$ & $0.009 \pm 0.001$ & $0.012 \pm 0.001$   \\
 & generated & $0.015 \pm 0.004$ & $0.009 \pm 0.002$ & $0.012 \pm 0.001$   \\

\bottomrule
\end{tabular}
\end{sc}
\end{small}
\end{center}
\vskip -0.1in
\end{table*}

\section{Implementation Details}\label{appendix:implementation-details}

\subsection{Data and Tokenization}

We use SMILES \citep{weiningerSMILESChemicalLanguage1988} based representations of molecules across all experiments. In all experiments we pre-train the \textsc{Joint Transformer} in an unsupervised manner using the ChEMBL database, a manually curated database of molecules with drug-like properties \citep{mendezChEMBLDirectDeposition2019}. As opposed to other datasets like ZINC \citep{irwinZINC20FreeUltralargeScale2020}, ChEMBL contains only molecules which have been synthesized. To ensure reproducibility and comparability with molecule generation baselines we use version $24$ of the database that contains $1.8$M compounds altogether and apply standard data processing used in the Guacamol benchmark \citep{brownGuacaMolBenchmarkingModels2019}. As for tokenization of the data, we use a tokenizer based on \citep{schwaller_probst_vaucher_nair_kreutter_laino_reymond_2020}. We additionally use an augmentation method of SMILES representations based on \citep{tetko2019augmentation} and similar to \citep{bagalMolGPTMolecularGeneration2022} across all experiments and methods. This ensures the transferability of results obtained by \cite{bagalMolGPTMolecularGeneration2022} to our experiments. 


\begin{figure}[h]\label{fig:joint-transformer-architecture}
\begin{center}
\makebox[\textwidth]{\includegraphics[width=0.8\paperwidth]{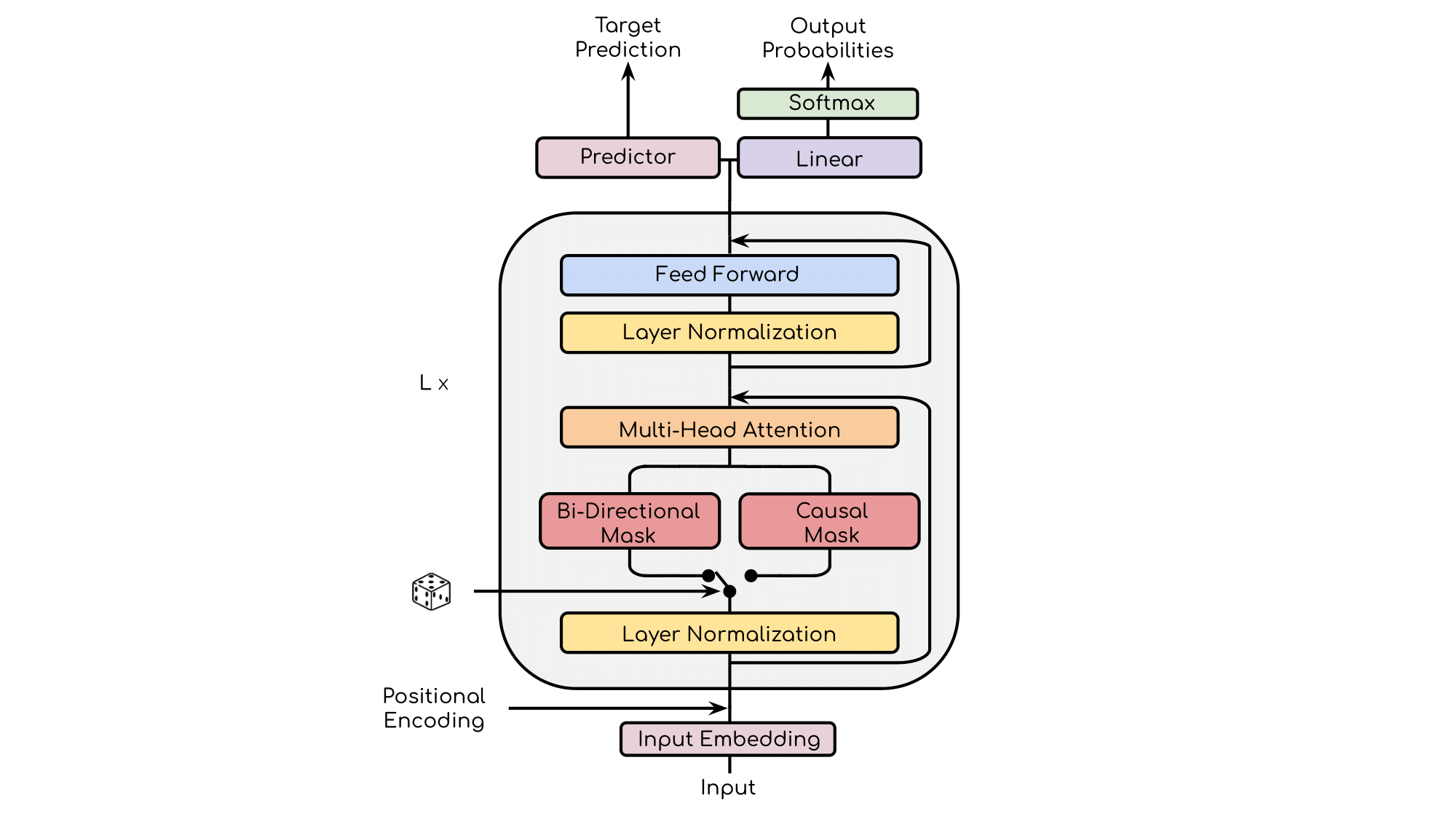}}
\end{center}
\caption{\textsc{Joint Transformer} architecture.}
\end{figure}

\subsection{Architecture}
Our implementation of the \textsc{Joint Transformer} follows the implementation provided by \citep{andrejMinGPT2023}, which is a re-implementation of a GPT-2 \citep{radford2019language} used by MolGPT \citep{bagalMolGPTMolecularGeneration2022}. The only difference is that during each forward pass, we switch between a causal and a bidirectional masking, depending on the task we are optimizing for. We additionally stack an MLP network on the top of the first output token for prediction. The complete list of hyperparameters is presented in Table~\ref{table:hyperparameters-model}. Our implementation results in a model with $6.5$M parameters.

\begin{table*}[h]
\caption{Model hyperparameters for the \textsc{Joint Transformer} used across all experiments.}
\label{table:hyperparameters-model}
\vskip 0.15in
\begin{center}
\begin{small}
\begin{sc}
\begin{tabular}{@{}lc@{}}
\toprule
Hyperparameter & Value \\
\midrule
activation fn & GELU \nocite{hendrycks2016gaussian} \\
embed dim & $256$ \\
num layers & $6$ \\
num heads & $8$ \\
feedforward dimension & $1024$ \\
feedforward bias & False \\
layer norm epsilon & $1\mathrm{e}{-5}$\\
predictor head & MLP \\
predictor num layers & 1 \\
predictor hidden dim & 100 \\
\bottomrule
\end{tabular}
\end{sc}
\end{small}
\end{center}
\vskip -0.1in
\end{table*}

\subsection{Training}\label{appendix:jt-training}

We provide the complete list of hyperparameters used for training \textsc{Joint Transformer} in Table~\ref{table:hyperparameters-train}. \textsc{Joint Transformer} was trained on a single 
NVIDIA GeForce RTX 2080 TI GPU for 4.2M iterations that took approximately seven days.

\begin{table*}
\caption{Training hyperparameters of the \textsc{Joint Transformer} used across all experiments.}
\label{table:hyperparameters-train}
\vskip 0.15in
\begin{center}
\begin{small}
\begin{sc}
\begin{tabular}{@{}lc@{}}
\toprule
Hyperparameter & Value \\
\midrule
batch size & $64$ \\
total number of training iterations & $4.2$ M \\ 
optimizer & AdamW \nocite{loshchilov2019decoupled} \\
weight decay & $1\mathrm{e}{-1}$ \\
beta 1 & $0.9$ \\
beta 2 & $0.95$ \\
maximum learning rate & $6\mathrm{e}{-4}$ \\  
minimum learning rate & $6\mathrm{e}{-5}$ \\ 
decay learning rate & True \\
warmup iterations & $2000$ \\
number of learning rate decay iterations & $4.2$ M \\
value to clip gradients at & $1.0$ \\
dropout & $0.1$ \\
task probability $p_{\rm{task}}$ & $0.95$ \\
\bottomrule
\end{tabular}
\end{sc}
\end{small}
\end{center}
\vskip -0.1in
\end{table*}

\subsection{Fine-tuning}

As \textsc{Joint Transformer} is a joint model, fine-tuning is achieved by standard training (Alg.~\ref{alg:joint-transformer-training}) on the supervised part of the dataset. Unless stated otherwise, we use the same set of hyperparameters for fine-tuning across all tasks, summarized in Table~\ref{table:hyper-fine-tune}. Fine-tuning on a single 
NVIDIA GeForce RTX 2080 TI GPU for 50K iterations takes approximately an hour. Hyperparameters not listed in Table~\ref{table:hyper-fine-tune} are shared with the pre-training task.

\begin{table*}[h]
\caption{Fine-tuning hyperparameters for the \textsc{Joint Transformer} used across all experiments.}
\label{table:molgen}\label{table:hyper-fine-tune}
\vskip 0.15in
\begin{center}
\begin{small}
\begin{sc}
\begin{tabular}{@{}lc@{}}
\toprule
Hyperparameter & Value \\
\midrule
decay lr & False \\
learning rate & $3\mathrm{e}{-5}$\\
num of iteration & 50K \\
task probability $p_{\rm{task}}$ & $0.1$ \\
\bottomrule
\end{tabular}
\end{sc}
\end{small}
\end{center}
\vskip -0.1in
\end{table*}

\end{document}